# A hierarchical semantic segmentation framework for computer vision-based bridge damage detection

Jingxiao Liu[1*], Yujie Wei[2], and Bingqing Chen[2]
*[1] Stanford University, Stanford, USA*
*[2] Carnegie Mellon University, Pittsburgh, USA*

## ABSTRACT

Computer vision-based damage detection using remote cameras and unmanned aerial vehicles (UAVs) enables efficient and low-cost bridge health monitoring that reduces labor costs and the needs for sensor installation and maintenance. By leveraging recent semantic image segmentation approaches, we are able to find regions of critical structural components and recognize damage at the pixel level using images as the only input. However, existing methods perform poorly when detecting small damages (e.g., cracks and exposed rebars) and thin objects with limited image samples, especially when the components of interest are highly imbalanced.

To this end, this paper introduces a semantic segmentation framework that imposes the hierarchical semantic relationship between component category and damage types. For example, certain concrete cracks only present on bridge columns and therefore the non-column region will be masked out when detecting such damages. In this way, the damage detection model could focus on learning features from possible damaged regions only and avoid the effects of other irrelevant regions. We also utilize multi-scale augmentation that provides views with different scales that preserves contextual information of each image without losing the ability of handling small and thin objects. Furthermore, the proposed framework employs important sampling that repeatedly samples images containing rare components (e.g., railway sleeper and exposed rebars) to provide more data samples, which addresses the imbalanced data challenge.

The paper evaluated the proposed framework with a public synthetic dataset - the Tokaido dataset. Our framework achieves a 0.836 mean intersection over union (IoU) for the structural component segmentation task and a 0.483 mean IoU for the damage segmentation task. Our results have in total 5% and 18% improvements for the structural component segmentation and damage segmentation tasks respectively, compared to the best baseline ensembled model. The experiment results validate the effectiveness of the proposed hierarchical semantic segmentation framework with important sampling and multi-scale augmentation.

## Introduction

Bridges are critical parts of transportation infrastructure that connect people, roadways, railways, communities, etc. Yet around 56,000 bridges in the U.S. are structurally deficient [1] which requires the development of efficient and accurate bridge health monitoring approaches.

Bridge conditions are currently monitored via manual inspection [2] and sensor instrumentation [3,4]. For the majority of bridges where sensors are not available, manual inspection by trained inspectors is the primary approach for SHM. However, manual inspection is labor-intensive, time-consuming, infrequent, and potentially dangerous [5]. Sensor-based bridge health monitoring methods overcome the drawbacks of manual inspection methods by continuously collecting structural performance data with sensing systems. However, such sensor-based methods are also costly and lack efficiency as they require on-site installation and maintenance of equipment and instruments [6].

Recently, computer vision-based (CV-based) techniques, which use remote cameras and UAVs to capture visual information of bridges, have gained great popularity in the bridge health monitoring domain. The low cost and high efficiency of CV-based techniques hold the potential to streamline the

---

[1*] Corresponding Author
*Email: liujx@stanford.edu*

current process of bridge health monitoring. A variety of computer vision techniques have been applied in the literature [3,7] to recognize structural elements and to detect bridge damages, e.g., concrete cracks and exposed rebar. In this work, we focus on detecting concrete cracks and exposed rebars on bridge columns.

Existing damage detection approaches could be broadly classified into two groups: object detection and semantic segmentation. Object detection-based approaches [8–10] treat damages, such as cracks, exposed rebars, and holes, as particular types of objects and aim at detecting their presence on images. The output of an object detection approach could be a bounding box or an outline surrounding the detected damage. Such object detection-based approaches are capable of detecting small objects on a high-resolution image, making them robust to view changes. However, object detection-based approaches could only output a rough outline of the damage, which is insufficient to support further analysis on the damage, such as evaluating the width, length, and growing speed of a crack. In contrast, semantic segmentation approaches [11–14] take an image as input and directly label each pixel based on its state (non-damage vs. different types of damages). Semantic segmentation-based approaches could provide a fine annotation of damages on an image but at the cost of less capability of detecting small and thin objects. In this study, semantic segmentation is adopted to further analysis of damages.

Applying existing semantic image segmentation methods to detect bridge damage is still challenging due to the following two reasons:

1) *Small and thin object challenge:* cracks and exposed rebars are small and thin objects without regular shapes. The pixel-level damage segmentation results are easily affected by other undamaged objects, making the pixel-level segmentation more difficult than traditional semantic segmentation tasks.
2) *Imbalanced data challenge:* Compared to regular components, damaged components only account for a very small part of all the collected bridge images, resulting in a highly imbalanced dataset.

To overcome the above challenges, this paper introduces a hierarchical semantic segmentation framework[1] for bridge damage segmentation. The framework contains two modules: 1) a component semantic segmentation module that predicts the category of bridge components (beams, columns, rails, sleepers, etc.) for each pixel using raw images as input and 2) a damage detection module that predicts the category of damage types (concreate cracks, exposed rebars, etc.) for each pixel using both raw images and the predicted component semantics as input. Instead of treating semantic information of structural components and damage separately, the proposed method leverages the domain knowledge that damage of interest could only reside on bridge columns and propagates the semantic information on structural components when detecting damages. Specifically, we first predict structural components and mask out non-column regions in the original images. In this way, the proposed method could partially address the *small and thin object challenge* by focusing only on the possible damaged regions using component semantics. During training and testing, we also conduct a multi-scale augmentation that resizes the original image with different ratios. This method provides the segmentation model with different zoom-in and zoom-out images, which shows closer views of small cracks and preserves contextual information of the entire structure. Furthermore, to address the *imbalanced data challenge*, we introduce an important sampling method for both the structural component and damage segmentation tasks. This method repeatedly samples images containing rare components (e.g., railway sleepers and exposed rebars) to make the dataset less imbalanced.

We evaluated the proposed framework using a publicly available synthetic dataset - the Tokaido dataset [15] - that includes images taken from 1,750 railway viaducts. We tested our framework for three tasks: 1) bridge component segmentation, 2) damage segmentation on extracted pure-texture images 3) damage segmentation on bridge images. In summary, our framework achieves a 0.836 mean IoU for the structural component segmentation task, a 0.712 mean IoU for the pure texture damage segmentation task, and a 0.483 mean IoU for the real scene damage segmentation task. The experiment results highlighted that the proposed method outperformed the best baseline ensembled models through semantic guided damage detection, importance sampling, and multi-scale augmentation.

---

[1] An implementation of the proposed framework can be found at https://github.com/jingxiaoliu/bridge-damage-segmentation

## Background

This section provides a review of existing computer-vision-based (CV-based) bridge damage detection methods and semantic image segmentation methods to identify research challenges and highlight the contributions of this study.

### *Computer-vision-based bridge crack detection*

CV-based techniques can automate part of the current laborious and costly process for manual vision inspection. The early work in CV-based bridge damage detection focused on extracting hand-crafted features defined based on engineering judgment. For instance, edge detection filters were applied to identify concrete cracking [16]. Given the tremendous success of CNN, it has become increasingly popular in recent years. Both structural component recognition [17] and damage detection [18] may be formulated as an object detection task, i.e., finding bounding boxes around the structural components/damage. In comparison, semantic segmentation is a promising approach that provides fine-granular, pixel-level information. Hoskere et al., [19] used three different neural networks to classify semantic information, presence of damage, and damage types, respectively. In this paper, our framework uses semantic information as input to the downstream damage segmentation task.

### *Semantic image segmentation*

Both the scene component recognition and damage detection tasks could be formed as a semantic segmentation problem. Semantic segmentation is the process of labeling each pixel on an image based on its semantics. From a high level of view, semantic segmentation is one of the most important tasks that paves the way to scene understanding. The following paragraphs provide an overview of existing semantic segmentation methods and cover the background needed to distinguish the proposed method from out-of-the-box machine learning models.

In general, semantic segmentation can be decomposed into two subtasks: feature extraction (encoder) and segmentation reconstruction (decoder). The encoder extracts features from an image to form a low-dimensional representation of integrated knowledge of the image content. Example encoders include convolutional neural networks (CNN), dilated convolutions, feature fusion, multi-scale prediction, recurrent neural networks, and transformers. The decoder takes an abstract feature vector as input and reconstructs a low-dimensional feature into a high-dimensional semantic segmentation image. Typical decoders include fully connected networks, feature pyramid networks, and object contextual representations [20–22].

In the proposed framework, we used a combination of three different types of encoders. Below is a detailed introduction to each of them.

1) HRNet [23]: Hi-Resolution Net is a widely convolutional neural network for extracting features for object recognition and semantic segmentation on high-resolution images. On the one hand, the model leverages conventional convolution layers to gradually convert high-resolution images into low-resolution features. On the other hand, the HRNet model leverages the multi-resolution connection to preserve detailed information that existed only in high-res textures.
2) ResNest [24]: ResNest is a representative feature extractor from the ResNet family that leverages residuals to address the vanishing gradients when training a deep neural network. Compared to the conventional ResNet, ResNest introduces the split-attention block to take advantage of channel-wise attention when extracting features while preserving the cross-channel feature correlations.
3) Swin [25]: Swin Transformer is an attention-based encoder that serves as a general-purpose backbone for semantic segmentation. As transformer-based methods gained huge success in the natural language processing domain, visual-transformers outperformed conventional CNN-based backbones on many public datasets. By introducing sliding windows in a hierarchical way, Swin Transformer could leverage the attention strategy when computing local features without introducing too much extra computation overhead.

The proposed method also uses two decoders as described below:

1) PSPNet [21]: The pyramid scene parsing decoder is known for its capability of extracting global context information for reconstructing semantic segmentation by using pyramid pooling.

Similar to image feature pyramids, PSP uses a pyramid of convolution layers that performs pooling on image crops at different scales. On the one hand, a large pooling layer allows the network to extract context information with a large perception field. On the other hand, a small pooling layer enables the use of local features for semantic understanding. In the proposed method, we used PSPNet together with ResNest and Resnet to create our models.

2) OCR [22]: Object-contextual representation is a decoder that aims at improving semantic segmentation accuracy by introducing soft object regions as part of the input of the network. Based on a rough semantic segmentation, the decoder first computes an object region representation, then classifies each pixel based on their relationships with the object region representations. The use of the intermediate object region representations enhances the concept of an "object" instead of labeling each pixel independently.

In the proposed framework, the final ensemble model comes from different combinations of the encoders and decoders above, including 1) HRNet + PSPNet; 2) ResNest + PSPNet; 3) HRNet + OCRNet; 4) ResNest+UperNet; and 5) Swin + UPerNet. The following section will provide a more detailed introduction to our segmentation framework.

## Hierarchical Semantic Segmentation Framework

This section will introduce the details of the proposed hierarchical semantic segmentation framework (as shown in Fig. 1.) that integrates the structural component segmentation and damage segmentation together to achieve accurate damage detection. The framework contains two modules: 1) a component semantic segmentation module that predicts the category of bridge components (beams, columns, rails, sleepers, etc.) for each pixel using raw images as input and 2) a damage detection module that predicts the category of damage types (concreate cracks, exposed rebars, etc.) for each pixel using both raw images and the predicted component semantics as input. Specifically, our implementation utilizes the semantic hierarchy between columns and column damages. It eliminates the influence of noncritical regions by masking out non-column pixels in the original images, forcing the damage segmentation model to focus only on the regions that have damages.

We first provide a summary of the proposed framework. In the training phase, images are input into the important sampling module and the structural component segmentation module that is learned following the flowchart in Fig. 2. The important sampling module repeatedly samples more images containing rare classes (e.g., exposed rebar) to address the imbalanced data challenge. The structural component segmentation module predicts component segments. Then, the importantly sampled data and column segment labels of each image are input to the next module that masks out non-column regions, which is to impose a hierarchical semantic relationship between columns and column damages. The masked images and corresponding labels are used to train the black-box image segmentation models in the ensemble learning module. During the testing phase, testing images are first input into the multi-scale augmentation module that augments the original images with different zoom ratios, which provides closer views of damages to overcome the small and thin objects challenge and retains contextual information of the entire structure. After masking out non-column components, we use the trained ensemble model to predict damage segmentation labels. The following subsections will provide more details of the proposed framework from five perspectives: 1) bridge component semantic segmentation; 2) importance sampling; 3) semantic-guided masking; 4) ensemble learning and 5) multi-scale augmentation.

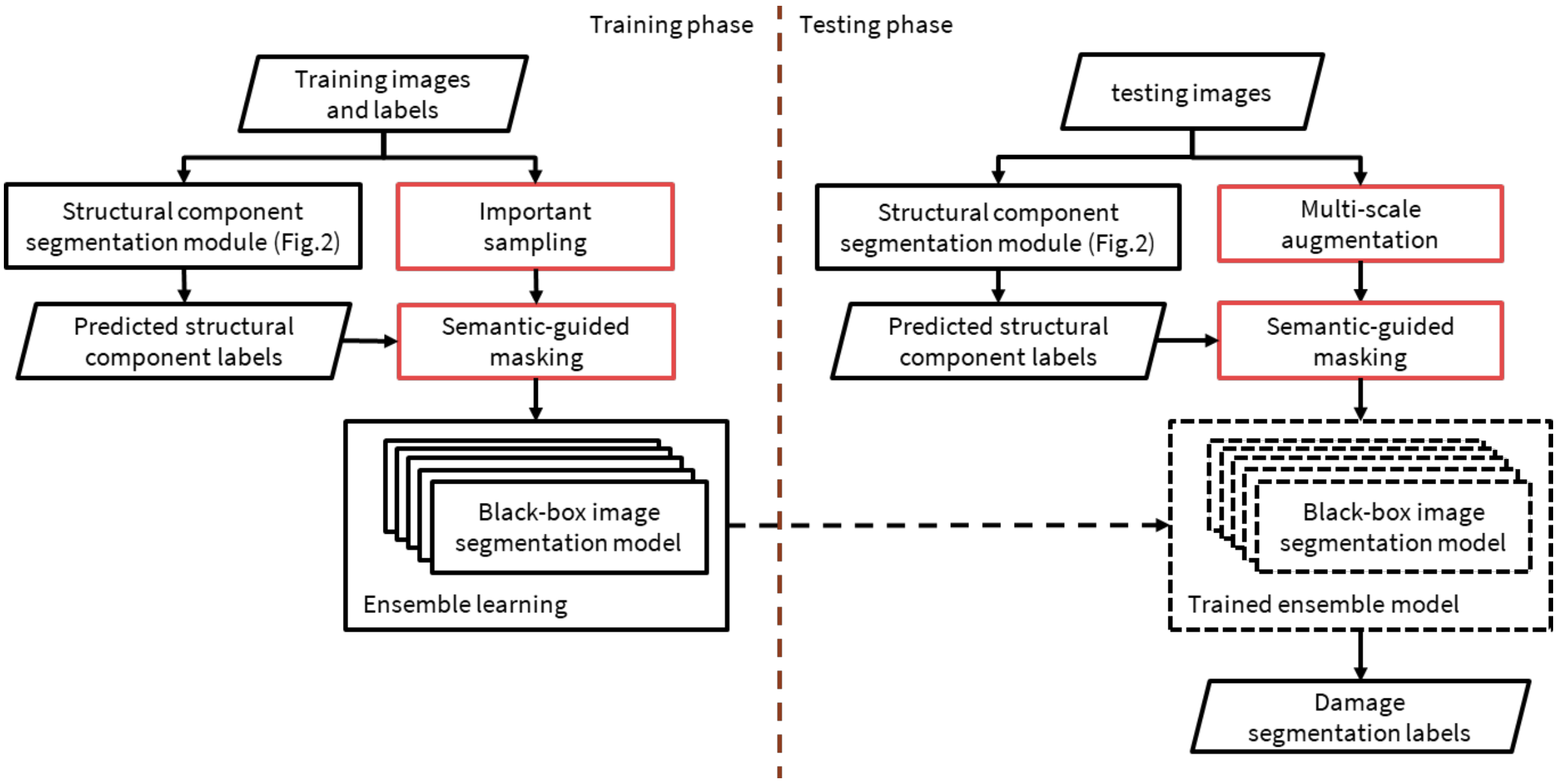

Fig. 1. Flowchart of our hierarchical semantic segmentation framework for bridge column crack detection.

***Bridge component semantic segmentation***

The workflow of bridge component semantic segmentation is shown in Fig. 2. During the training phase, bridge component images together with their semantic segmentation (Non-bridge, Slab, Beam, Column, Nonstructural, Rail, Sleeper, Other) will be used for training a set of deep learning models. During the testing phase, the trained model could take a bridge scene image as input and predict its semantic segmentation. Notice that the workflow also includes important sampling, multi-scale augmentation, and ensemble learning. The mechanisms of these three features, which will be described in the following subsections, are in general the same as those in the damage detection framework shown in Fig. 1. Specifically, for the bridge component semantic segmentation task, the important sampling is conducted on railway components (e.g., rails and sleepers) as they are more difficult to recognize due to their thin shape and rare occurrence in the dataset.

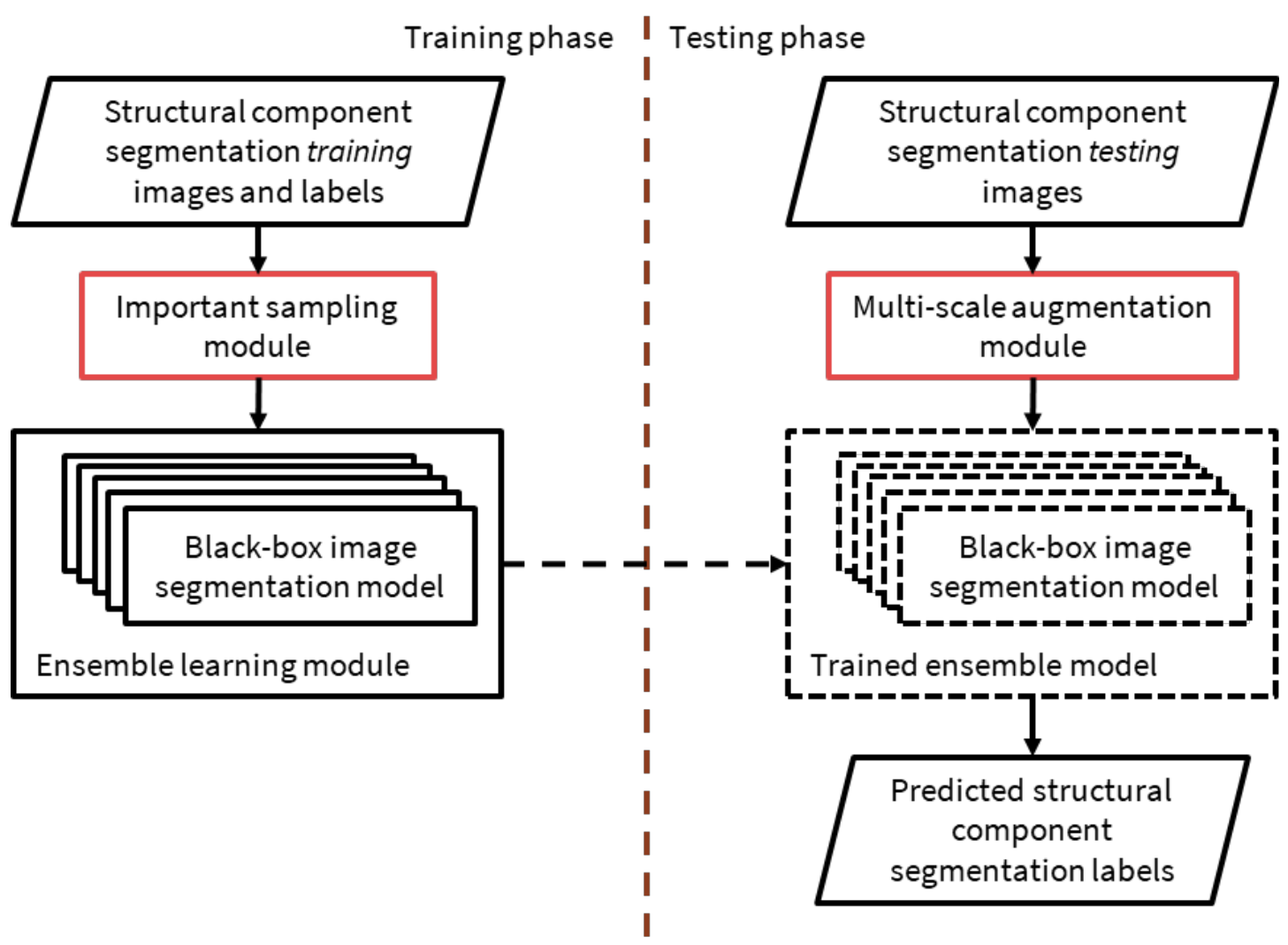

Fig. 2. Flowchart of our structural component segmentation module.

***Important sampling***

As mentioned above, importance sampling is employed to repeatedly samples images containing components of interest to overcome the imbalanced data challenge. The importance weights are calculated based on class occurrence, i.e., the less a particular class of component presents in the dataset, the higher importance weight it has during resampling. For example, in the implementation the sleeper class was assigned a much higher importance weight because only around 6% of images have sleeper components in the training dataset. The importance sampling algorithm contains two hyper-parameters: 1) $n_m$, the minimum number of pixels of the rare class in the image, and 2) $r$, the number of repeating. We first count the number of pixels of the important component in each image defined as $n$. Then, we repeated sample $r$ times if $n > n_m$. The two hyper-parameters for important sampling are selected through fine-tuning in the evaluation step.

***Semantic-guided masking***

Semantic-guided masking bridges the bridge component semantic segmentation module and the damage detection module. Given an image as input, the bridge component recognition module first computes the semantic segmentation of bridge components. The raw image together with the semantic segmentation will be passed to the damage detection module. Since damages could only reside on bridge columns, the damage detection module leverages semantic information to mask out pixels that do not belong to column segments in each image, which allows the model to focus on distinguishing damage vs non-damage regions. Specifically, the semantic-guided masking replaces the pixels that are non-column with black RGB values (i.e., [0, 0, 0]). Fig. 3. shows an example of the original image and the masked image.

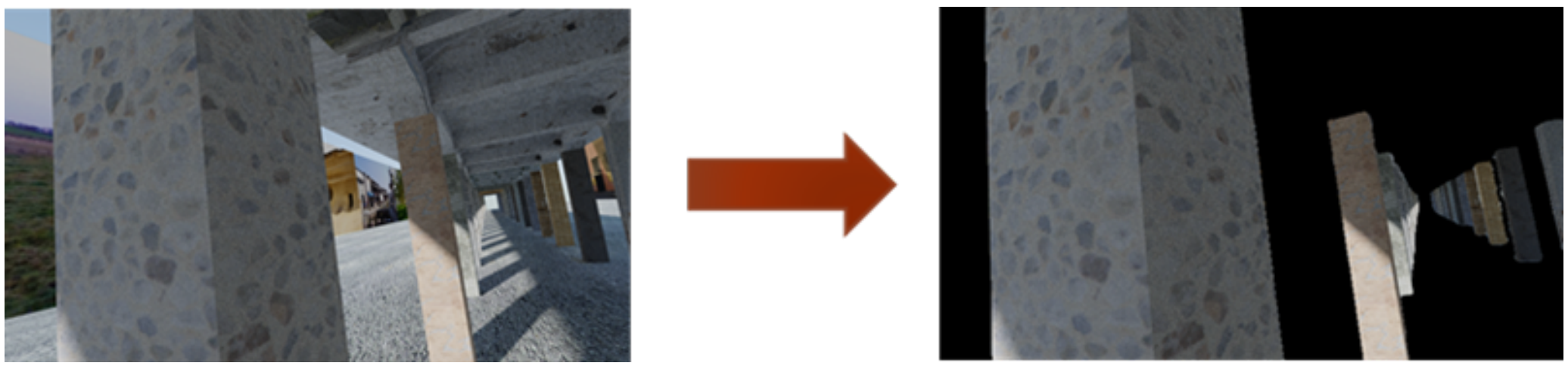

Fig. 3. An example of our semantic-guided masking module. Non-column regions are replaced with black.

***Ensemble learning***

In order to reduce the variance due to model training and data sampling, we use a simple majority vote as a way to ensemble machine learning models for both baseline approaches and the proposed method in the implementations. Specifically, we conduct majority voting over predictions using multiple image segmentation methods in order to improve the final prediction accuracy. For both the baseline approach and the proposed method, we first trained five image semantic segmentation models indivisually (HRNet + PSPNet, ResNest + PSPNet, HRNet + OCRNet, ResNest+UperNet, and Swin + UPerNet) as introduced in the background section. Then, the final prediction of each pixel is the mode of the predictions of the five segmentation models.

***Multi-scale augmentation***

Despite that the data has been augmented during the training phase (flip, crop, scale, photo-metric distortion, etc.), the semantic segmentation and damage detection modules are still susceptible to scale changes during the testing phase. Therefore, multi-scale augmentation is also employed in the testing phase is to make multiple predictions of the same region of an image using crops with different scales as input. Our segmentation models are trained and tested with the same size of image patches cropped from each input image. By resizing the original image with different ratios and sliding the crop window across the image, the segmentation model could make predictions using both detailed textures of thin objects (e.g., railway sleepers and damages) and preserves context information of components and surrounding objects. Fig. 4 shows an example of the test-time multi-scale augmentation.

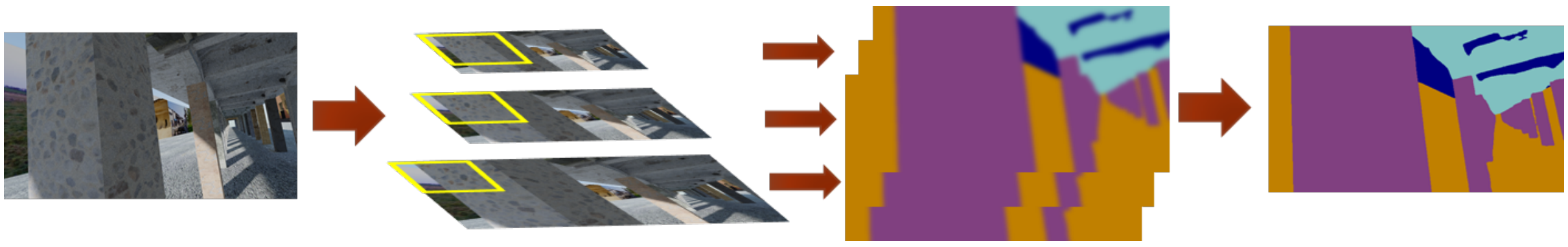

Fig. 4. An example of the test-time multi-scale augmentation module. The final prediction ensembles predictions of the same image with multiple zoom ratios.

## Evaluation

In this section, we evaluated the proposed approach for structural component semantic segmentation and damage segmentation using the Tokaido dataset [13]. Below first describes the dataset used for evaluation and the implementation details of the proposed approach, followed by evaluation results on the Tokaido dataset and the discussions on the results.

### *Dataset description*

The Tokaido dataset is a publicly available synthetic image dataset that consists of 1,750 railway viaducts with random geometry realized by the actual design procedure. In the dataset, random damages, including concrete cracks and exposed rebar, could present on the viaduct columns.

The dataset supports evaluations from three perspectives: 1) structural components segmentation, 2) damage segmentation for real scene images, and 3) damage segmentation for pure texture images. For the structural component segmentation task, there are 8,648 images (as shown in Fig. 5) with seven classed of components: non-bridge, slab, beam, column, non-structural, rail, and sleeper. For the damage segmentation tasks, the dataset has 7,990 real scene images (as shown in Fig. 6) and 2,700 pure texture images (as shown in Fig. 7) with three classes: non-damage, concrete damage, and exposed rebar.

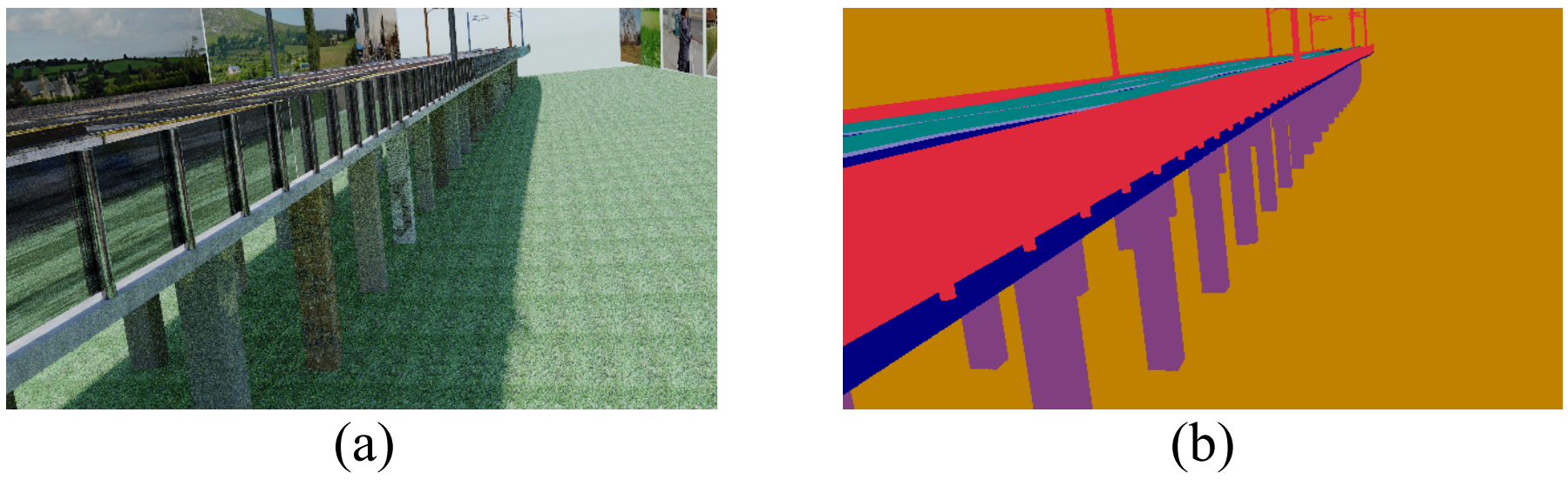

(a) (b)

Fig. 5. An example of (a) synthetic image for structural components segmentation and (b) ground truth structural components labels.

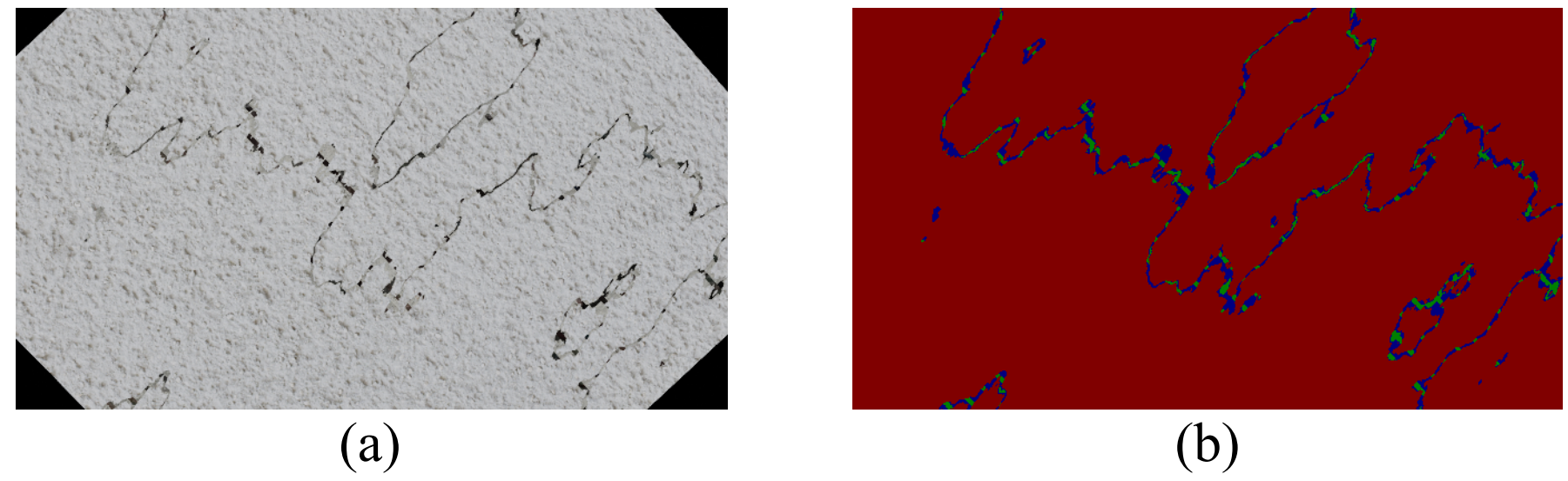

(a) (b)

Fig. 6. An example of (a) synthetic image for pure texture damage segmentation and (b) ground truth damage labels.

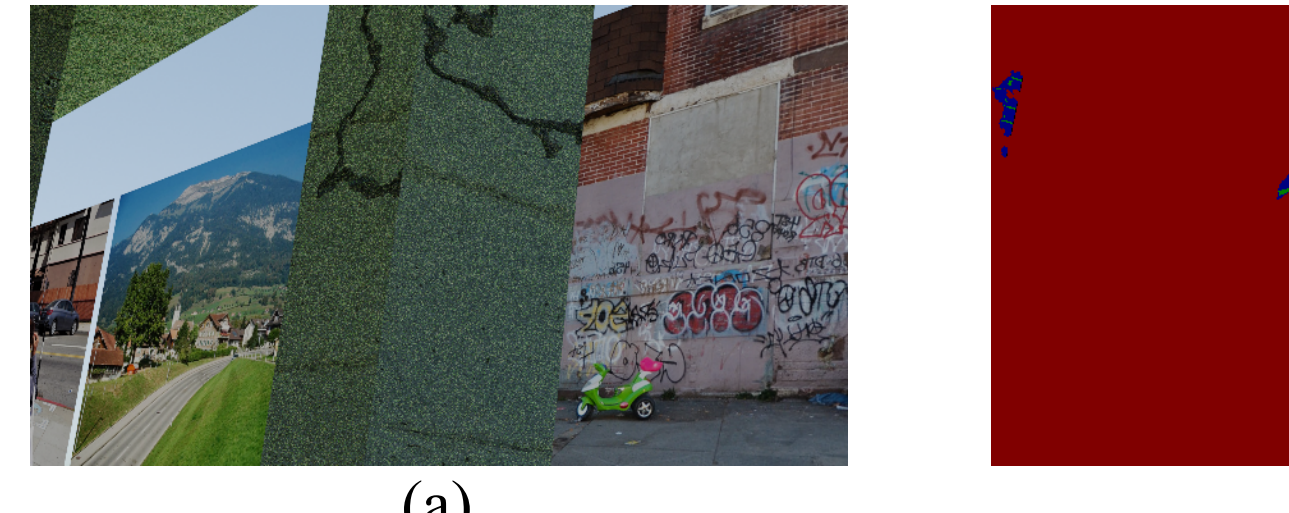

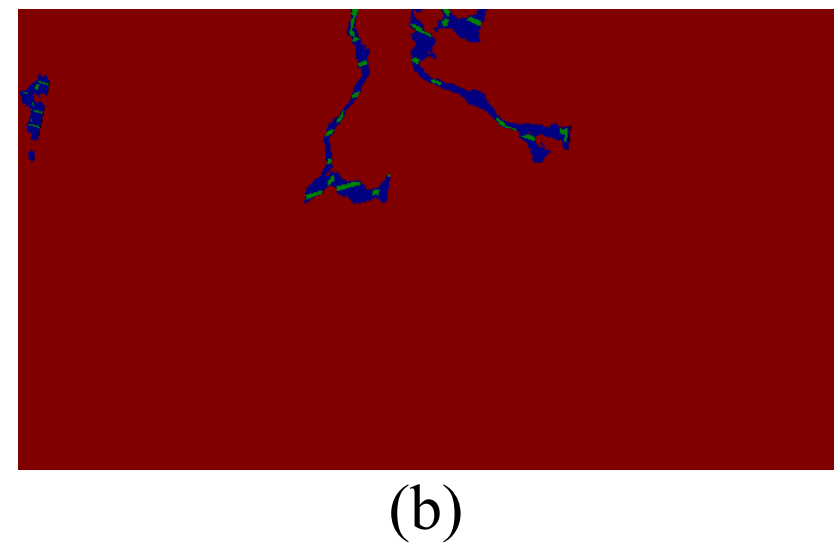

(a) (b)

Fig. 7. An example of (a) synthetic image for real scene damage segmentation and (b) ground truth damage labels.

***Implementation***

This subsection introduces the setup of our hierarchical semantic segmentation framework. Specifically, to overcome the imbalanced data challenge, we first conduct important sampling by repeatedly sampling images containing railway components and exposed rebar damages ten times. Then, we split the sampled dataset into a 90% training set and a 10% validation set. One should note that the splits for structural components and real scene damage segmentation tasks are based on viaducts, i.e., we split viaducts geometry into two random subsets in a ratio of 9:1 and use images taken from training viaducts subset for model training and others for model validation. Furthermore, each semantic segmentation model is trained with two 16 GB Nvidia Tesla P100 GPUs and optimized using a stochastic gradient descent optimizer with a polynomial learning rate schedule, in which the learning rate is decayed to a polynomial function. Hyper-parameters, such as learning rate, batch size, crop size, and important sampling parameters, are selected via fine-tuning and cross-validation. The implementation details could be found at the code repository.

***Results and discussion***

This subsection presents our evaluation results and discussions of the findings. Intersection over Union (IoU) is used as the primary metric to evaluate the model performance. In particular, IoU is computed by dividing the area of overlapping regions between the predicted segmentation and the ground truth segmentation by the area of union. If the predicted segmentation aligns with the ground truth perfectly, IoU should be 1. When the predicted segmentation has no overlapping region with the ground truth segmentation, the IoU will be 0.

Fig. 8 shows examples of the structural component and damage segmentation results predicted using our framework. Compared to the corresponding ground truth labels in Fig. 5-7, we can observe that our framework successfully predicts structural components and damaged regions.

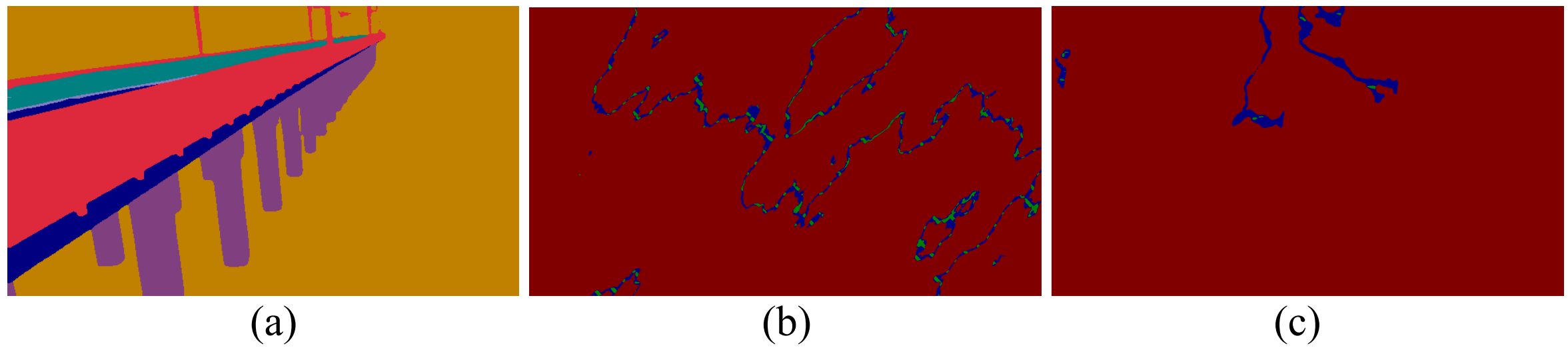

(a) (b) (c)

Fig. 8. (a) Structural component and (b), (c) damage segmentation results for the examples in Fig 4-6.

Table 1-3 summarize the IoU results for structural component segmentation, damage segmentation of pure texture images, and damage segmentation of real scene images, respectively. For all the three segmentation tasks, the proposed approach (ENS+I.S.+M.S. and ENS+I.S.+M.S.+S.G.M.) outperformed baseline approaches that do not use the important sampling and/or multi-scale testing. Specifically, for the structural component segmentation task, the important sampling (I.S.) and multi-scale testing (M.S.) improved the mean IoU by 4% and 1%, respectively; for the damage segmentation of pure texture tasks, I.S. and M.S. improved the mean IoU by 25% and 0.1%, respectively; and for the damage segmentation of real scene images task, I.S. and M.S. improved the mean IoU by 15% and 2%, respectively. In addition, for damage segmentation of real scene images, our framework that conducts

semantic-guided masking (S.G.M.) improved the damage detection results by 1.6%.

Table 1. IoUs for structural component segmentation. ENS stands for the ensembled model. I.S. and M.S. mean training with important sampling and testing with multi-scale, respectively.

| | Slab | Beam | Column | Non-structural | Rail | Sleeper | Average |
|---|---|---|---|---|---|---|---|
| ENS | 0.891 | 0.880 | 0.859 | 0.660 | 0.623 | 0.701 | 0.785 |
| ENS+I.S. | 0.915 | 0.912 | 0.958 | 0.669 | 0.618 | 0.892 | 0.827 |
| ENS+I.S.+M.S. (ours) | **0.924** | **0.929** | **0.965** | **0.681** | **0.621** | **0.894** | **0.836** |

Table 2. IoUs for damage segmentation of pure texture images.

| | Concrete damage | Exposed rebar | Average |
|---|---|---|---|
| ENS | 0.356 | 0.536 | 0.446 |
| ENS+I.S. | **0.708** | 0.714 | 0.711 |
| ENS+I.S.+M.S. (ours) | 0.698 | **0.727** | **0.712** |

Table 3. IoUs for damage segmentation of real scene images. S.G.M. means semantic-guided masking.

| | Concrete damage | Exposed rebar | Average |
|---|---|---|---|
| ENS | 0.235 | 0.365 | 0.300 |
| ENS+I.S. | 0.340 | 0.557 | 0.448 |
| ENS+I.S.+M.S. | 0.350 | 0.583 | 0.467 |
| ENS+I.S.+M.S.+S.G.M. (ours) | **0.379** | **0.587** | **0.483** |

The result tables show that the important sampling method has the largest improvement (around a 20% increase of IoU) on segmenting railway sleepers and exposed rebar damages. These improvements show the effectiveness of important sampling on recognizing classes that have less occurrence and smaller areas. Furthermore, the IoU results for predicting non-structural, rail components, and damage segments increase after conducting multi-scale testing, which validates the effectiveness of our method for accurately predicting thin objects.

**Conclusions and Future Work**

In this work, we used a synthetic dataset generated from computer graphics algorithms to train a semantic segmentation model for understanding infrastructure scenes and detecting damage regions. The proposed framework addressed three key challenges when dealing with infrastructure scenes. First, the proposed method improved scene understanding and damage detection accuracy by resampling components of interest that are rarely presented in images. The experiment results highlighted that the resampling technique addressed the category imbalance problem existing both across different data points and within a particular data point. Second, the proposed method handles the perspective view changes using testing time multi-scale augmentation. Considering that objects of interest could have different scales in the captured images, the proposed method recognizes scene components and damages using crops with different scales at testing time to mitigate the effect of viewpoints. Third, the proposed damage detection module leverages the domain knowledge that concrete damages could only be located on concrete components to further refine the damage detection results. By masking the non-concrete regions when feeding an image to the damage detection network, the proposed method could use the predicted semantics to guide the damage detection, thus improving the prediction accuracy and reducing the computation time. The experiment results discussed in the previous section have validated the feasibility and the effectiveness of the proposed framework.

Notice that in this study it is assumed that semantic segmentation labels are available for the training set. Given the practical difficulty of collecting and annotating a sufficiently large dataset for deep learning from real-world scenes, we envision using either sim-to-real transfer learning from synthetically generated data or self-supervised learning on unlabeled data as the pathway for our current work to be transferable to practical application.